%% file: template.tex
\documentclass{Interspeech2024}

\title{Word Sense Disambiguation in Native Spanish: \\ A Comprehensive Lexical Evaluation Resource}

\name{Pablo Ortega, Jordi Luque, Luis Lamiable, Rodrigo López, Richard Benjamins} 

\address{Telef\'onica Innovación Digital, Research, Spain\\
         }

\email{{jordi.luque}@telefonica.com}

\keywords{word sense disambiguation, word sense discrimination, WSD, lexical semantics, sense dataset, sense inventory}
    
\usepackage{color}

\usepackage{hyperref}
\usepackage{multibib}
\usepackage{multirow} 

\newcites{languageresource}{~}

\usepackage{graphicx}
\usepackage{tabularx}
\usepackage{soul}
\usepackage{comment}
\usepackage{nicematrix}
\usepackage{booktabs}
\usepackage[utf8]{inputenc}
\DeclareUnicodeCharacter{2212}{-}

\usepackage{epstopdf}
\usepackage[utf8]{inputenc}

\usepackage{xcolor}
\usepackage{hyperref}
 \definecolor{darkblue}{rgb}{0, 0, 0.5}
  \hypersetup{colorlinks=true, citecolor=darkblue, linkcolor=darkblue, urlcolor=darkblue}

\usepackage{xstring}

\begin{document}
\maketitle
\begin{abstract}

Human language, while aimed at conveying meaning, inherently carries ambiguity. It poses challenges for speech and language processing, but also serves crucial communicative functions. Efficiently solve ambiguity is both a desired and a necessary characteristic. The lexical meaning of a word in context can be determined automatically by Word Sense Disambiguation (WSD) algorithms that rely on external knowledge often limited and biased toward English. When adapting content to other languages, automated translations are frequently inaccurate and a high degree of expert human validation is necessary to ensure both accuracy and understanding. The current study addresses previous limitations by introducing a new resource for Spanish WSD. It includes a sense inventory and a lexical dataset sourced from the Diccionario de la Lengua Española which is maintained by the Real Academia Española. We also review current resources for Spanish and report metrics on them by a state-of-the-art system.
\end{abstract}

\section{Introduction}

The goal of language is the communication of meaning but the human language is inherently ambiguous, with uncertainty brought on by pragmatic, syntactic and lexical factors. A term becomes lexically ambiguous when it has several meanings. For instance, the Spanish term "banco" can be used to describe both a financial organization and a bench in a public space. If the intended meaning cannot be determined from the context, this form of ambiguity may cause misunderstanding and misinterpretation. In the field of Natural Language Processing (NLP), the difficult task of computationally determining the appropriate meaning of a word in a particular context is known as word sense disambiguation (WSD). The ability to navigate and resolve ambiguity is crucial for successful computational systems. 
Despite the intended lexical meaning of a word in context can be, to some extend, determined automatically by WSD algorithms, unfortunately, external knowledge like a predefined sense inventory is a fundamental component for these algorithms. It provides the essential data to associate senses with words, and usually is a scarce resource, mainly in English, an issue known as the knownledge acquisition bottleneck \cite{ijcai2018p812}.

The NLP community tends to maintain the working assumption that word meaning can be discretized in a finite number of classes \cite{el-sheikh-etal-2021-integrating}, thus casting polysemy resolution as a multi-class classification problem, where the classes, e.g. the senses are specific to a word. Senses are registered in a dictionary like resource called the sense inventory. In WSD the sense inventory is virtually always The Princeton WordNet (WNG)~\cite{ijcai2021p593} together with some parallel corpus annotated with senses, like MultiSemCor and Babel, just multilingual versions of WordNet. Nonetheless they suffer by the use of semi-automatic methods for data harvesting~\cite{taghipour-ng-2015-one}, non-accurate automatic translations or by the semi-automatic validation of the word senses and granularity~\cite{Lacerra_Bevilacqua_Pasini_Navigli_2020}. A drawback stemming from the fact that every language is inherently subject to change and interpretation which undoubtedly requires a native speaker validation and a high level of expertise.

However, as mentioned in a previous study \cite{ijcai2021p593}, all of these inventories encounter a challenge known as the fine-granularity problem. This issue arises when distinguishing between various meanings of the same word becomes challenging, even for humans. For instance, WordNet lists 29 senses for the noun {\it line}, including two that differentiate between a set of horizontally laid out things and one laid out vertically. To address the excessive granularity, coarser-grained inventories have been suggested, but mainly performed on the English language with few extensions to other main languages, such as French, German, Italian and Spanish. Moreover, the meanings of non-English words are translated from English, ignoring many of the nuances of different senses in specific situations.

The Spanish language has several specific characteristics that justifies a deeper look in relation to WSD. It is spoken by almost 600 million\footnote{{\url{https://www.ethnologue.com/language/spa/}}} people in the world of which about 100 million are non-native speakers. A distinguishing factor of the Spanish language, compared to most other languages, is that it is a globally “regulated” language with respect for local variations. The Spanish Royal Language Academy (Real Academia Española, RAE 
\footnote{\url{https://www.rae.es/}}) 
is a 300-year-old Institution that, in collaboration with all local Spanish language academies, is actively monitoring and managing the Spanish language in all its geographical regions. The main materialisation of this work are the official dictionary of Spanish language  DLE
, with more than 94,000 entries, 30\% more than other commercial Spanish dictionaries, and an average of 2.5 meanings per entry and the Student´s Dictionary in Spanish (SDS) a lexicographic work specially designed for students of Spanish. The wide usage of Spanish across the world, combined with the normative approach for its evolution, clearly justifies a Spanish-specific approach for WSD. 

Among the contributions in this work, we provide a new lexicon resource for WSD in native-Spanish. Furthermore, we provide a comprehensive review of the existing resources and approaches to Spanish WSD and a specific approach by fine-tuning  BERT and RoBERTa based models, by using different combinations of Spanish WSD resources. Finally, we report the performance metrics on the most popular benchmarks and on the new evaluation resource in native Spanish. The evalutation dataset\footnote{{Removed due to anonymisation reasons.}}
 and the final systems are provided, as a resource publicly accessible in Huggingface\footnote{Removed due to anonymisation reasons.}. The models' performance is able to either achieve or surpass state-of-the-art results attained by most of the supervised neural approaches on the Spanish WSD task. This demonstrates the advantages of incorporating lexical knowledge, specifically expert validated senses and glosses, in fine-tuning neural models for WSD.

\section{Related Work}
\label{related}
Early WSD experiments included manually crafted lexicons and rules~\cite{10.1145/1459352.1459355}. But as this field's study has advanced, more complex answers have been put forth. There are different approaches in the literature to tackle WSD, including supervised methods, unsupervised methods and knowledge or graph-based methods~\cite{ijcai2021p593}. 

Supervised methods use labeled training data, e.g., sets of encoded examples together with their sense classes or labels, to learn a classifier that can predict the correct meaning of a word in context assigning, e.g., by maximizing the similarity to a single meaning like a single-label classification
problem or as a soft multi-label classification problem in which multiple senses can be assigned to each target word~\cite{conia-navigli-2021-framing}. On the other hand, unsupervised methods are based on unlabeled corpora and do not exploit any manually sense-tagged corpus to assign senses to words in context, e.g., based on their co-occurrence patterns \cite{10.1145/1459352.1459355}. Finally, knowledge-based methods \cite{10.1162/COLI_a_00164}, assign senses to words based on their semantic relationships, relying on external resources such as Machine-readable dictionaries (MRDs), like as WordNet \cite{ijcai2021p593}, or other structured or non-structured knowledge sources like Sense-Annotated Corpora, SemCor~\cite{miller-etal-1993-semantic}, among others. 
Supervised systems, particularly those based on Transformer-based language models, have become the standard approach for WSD \cite{barba-etal-2021-consec}. These systems leverage large-scale pre-trained models to learn contextual representations of words and their senses, which can then be used for disambiguation.

Although there is no model exclusively developed for Spanish, there are several multilingual models capable of disambiguating Spanish texts. Among these models we found mainly both supervised and knowledge-based methods. In the supervised category, we encountered systems such as AMuSE-WSD \cite{orlando-etal-2021-amuse} or Multimirror \cite{ijcai2021p539}. The former was the first approach to frame the WSD task as down-stream task, using multi-class classification and offering a multilingual WSD system built on top of modern pre-trained language model. In contrast, Multimirror~\cite{ijcai2021p539} is an approach that addresses the data scarcity issue in a target language by proposing a cross-lingual sense projection approach. It involves aligning parallel sentences from English to a target language for augmenting the low-resourced language.

On the other hand, we also encountered a few multilingual knowledge-based systems. First among them is SensEmBERT \cite{Scarlini_Pasini_Navigli_2020}, where the authors create latent representations of word meanings and glosses in various languages, by using sentence embeddings from a multilingual mBERT model and comparing distances with candidate meanings. Additionally, there is the work reported in \cite{maru-etal-2019-syntagnet}, that introduces SyntagNet a resource made up of manually disambiguated lexical-semantic combinations. In their experiments they train UKB, a graph based system~\cite{10.1162/COLI_a_00164}, achieving remarkable performance when combining with various Lexical Knowledge Bases (LKBs). 
Finally, we found mBERT-UNI~\cite{su-etal-2022-multilingual}, a supervised model and knowledge-based. It is a supervised framework incorporating a lexical unified representation space for multilingual WSD. 

\section{Resources for WSD}
The effective execution of WSD algorithms heavily rely on the availability and quality of resources like LKBs. In this section, we delve into the most popular resources that have driven the WSD task.

\subsection{Sense Inventories}
\label{sec:senseinventories}

The sense inventories list the various meanings a word may have. The most popular are: 
\begin{itemize}
    \item \textbf{Wordnet} \cite{miller-1992-wordnet} is a large lexical database with over 120k concepts that are related by over 25 types of semantic relations and comprise over 155k words (lemmas), from the categories Noun, Verb, Adjective and Adverb. It organizes concepts in synsets -- sets of synonyms -- and provides, for each of them, one or more sentences in which each one is used with each meaning.
\item \textbf{Babelnet} \cite{NAVIGLI2012217} is a multilingual semantic network  with the principal objective of functioning as an extensive "encyclopedic dictionary". Its initial iteration was primarily focused on automatically linking Wikipedia articles and WordNet senses. However, the current iteration, BabelNet 5.0, has considerably expanded its knowledge acquisition scope, drawing insights from 51 distinct sources.

\end{itemize}

\subsection{Sense-Annotated Data}

 The resources for WSD in languages other than English are much more limited. We focus primarily on the most prominent multilingual datasets and present two new Spanish resources.
 
\subsubsection{Training Data}
\label{sec:trainingdata}
\begin{itemize}
    \item \textbf{SemCor} \cite{miller-etal-1993-semantic} is the largest manually sense-annotated English corpora, is a significant resource for training supervised Word Sense Disambiguation systems. It consists of over 226,000 sense annotations across 352 documents. 
    \item \textbf{MuLaN} \cite{ijcai2020p531}is a specialized tool for WSD, which can automatically create sense-tagged training datasets in multiple languages. What distinguishes MuLaN is its extensive range of sense categories, achieved through its integration with the BabelNet inventory, in contrast to SemCor, which exclusively depends on WordNet inventory. The dataset
    includes translations in four distinct languages: German, Spanish, French, and Italian. 

    \item \textbf{SDS}\footnote{\url{https://www.rae.es/obras-academicas/diccionarios/diccionario-del-estudiante}} the Student´s Dictionary in Spanish is a lexicographic work specially designed for students of Spanish. It is composed of around 40K terms and is elaborated taking into account text and reference books employed in the educational systems of Spain and America. This resource is available in both printed form and as a mobile application.
    \end{itemize}

\subsubsection{Evaluation Data}
\label{subsec:evaldata}
\input{table_semevals}

\begin{itemize}
\item \textbf{SemEval-2013 task 12} \cite{semeval-2013}: The test set was made up of 13 articles that were drawn from datasets from the workshops on Statistical Machine Translation. The articles were available in 4 different languages, in which we can find Spanish. The process of annotation was successfully carried out by a pair of native speakers. 

\item \textbf{SemEval-2015 task 13} \cite{semeval-2015}: Is a dataset that encompasses both types of inventories, including named entities and word senses, within distinct domains such as biomedicine, mathematics and computer science, as well as a broader domain-focused on social issues. This dataset was meticulously developed in three languages, namely English, Italian, and Spanish, using parallel texts. The annotations were carried out independently and manually by various native or fluent speakers.

\input{table_comparison_eval}

\item{\textbf{DLE}\footnote{\url{https://dle.rae.es/}}}: The Diccionario de la Lengua Española, holds significant value as the fundamental reference for defining, spelling, grammar and proper usage of Spanish words. The creation of the DLE is the result of a collaborative effort among various academies worldwide\footnote{\url{ https://www.asale.org/} }, including those from Spain, North America, Equatorial Guinea, Filipinas and  South America and Central America.

Its broad scope encompasses the lexicon commonly utilized in Spain and across Spanish-speaking nations

\end{itemize} 

\vspace{-0.25cm}
\section{Experiments and Results}
\label{exp_setup}

\subsection{Construction of the training and evaluation datasets}
\input{table_dle_example}
We use the Spanish versions of the SE13 and SE15 evaluation datasets from the {\it mwsd-datasets}~\footnote {\url{https://github.com/SapienzaNLP/mwsd-datasets}} repository that are extracted from the Babelnet and Wordnet inventories, see section \ref{sec:senseinventories}, 
For this work, we decide to employ the {\it wn} split, which includes a subset of those instances tagged with a BabelNet synset that also contains a sense from the WordNet. The {\it wn} split is used for both evaluation datasets. In the table \ref{table:table_semevals} we report the number of instances, comprising a total of 1260 and 1043 samples for the SE13 and SE15, respectively.

To create the DLEval evaluation resource we collected, by a web crawling, a subset of the official DLE dictionary (see section \ref{subsec:evaldata}). It is comprised of $m$ more than 12K target words $w$, the set of meanings $S_{w}$ for each $w$, $S_{w} = {s_{w1}, s_{w2}, \dots, s_{wk}}$, and along with its corresponding set of glosses $G_w = {g_{w1}, g_{w2}, \dots, g_{wk}}$, which adds the contextualized information. Every gloss demonstrates the precise interpretation of a target word, providing a sentence example for its use and contextualizing a concrete meaning.  

Finally, for building a testing example for classification, we compose it by the lemma of the target word, the lemmatized target gloss along with four different senses, including the target word sense.
Following the standard in the WSD task, we generate a XML file containing all lemmatized glosses annotated with the Part of Speech (PoS) information, alongside a gold file that maps the {\it sense} of each word to the target lemma within 
corresponding gloss. For the PoS tagging and word lemmatization we use the Freeling~\cite{padro-stanilovsky-2012-freeling} tool.

Table \ref{table:table_comparison_eval} reports on the number of extracted instances. In order to promote transparency, reproducibility and the comparison with previous and future WSD apporaches, we will publicly release a small portion of the DLEeval resource together with the models and the python code upon acceptance of this work. The full DLEeval employed in this work, consists of around 12,000 samples, being all instances  single-words(SW). It represents around a 20\% of the total number of lemmas in the official DLE dictionary. Nonetheless, this number represents almost ten times compared to the instances available in SE13 or SE15.

For the MULAN and SemCor training datasets we use the Spanish versions of MuLaN and SemCor provided by \cite{su-etal-2022-multilingual}, containing the target definition or sense, the context and the word translated into Spanish. For the completion of the classification instances, we employ Wordnet to add three additional senses. These definitions, originally in English, are translated using the Google translation tool \cite{wu2016googles}. 

For the SDS training set a snapshot of the dictionary, provided by the RAE, was processed following the same approach as for the DLEeval dataset construction. In this case, given the dictionary’s extensive word count of approximately 40,000 words, we were able to extract a total of 46,000 different instances with its corresponding definition and examples of use. In addition to the three original datasets, MULAN, SemCor(SC) and SDS, we generated three different combinations of them aiming to perform an ablation study of their importance depending on the evaluation data.

\input{table_all_in_one}
\vspace{-0.25cm}
\subsection{Fine-tuning Large Language Models}

We adopt a similar approach as in \cite{yap-etal-2020-adapting}, treating the task as a multi-label classification problem. In this setting, the WSD task consists on disambiguate the senses of a target word $w_t$ in a sentence $W = {w_1, \dots, w_t, \dots, w_m}$. For each target word $w_t$, the goal is to map it to a pre-defined sense $s \in S_{wk}$, where $S_{kw} = {s_{1w}, s_{2w}, \dots, s_{kw}}$ is the set of $k$ pre-defined candidate senses for $w$. The meaning of each sense is defined by the gloss. The candidate senses have a corresponding gloss set defined as $G_{wk}= {g_{w1}, g_{w2}, \dots, g_{wk}}$. The models used in our experiments are BERT and RoBERTa based models and fully pre-trained using Spanish corpus. BETO~\cite{CaneteCFP2020} is a BERT-base model with 110M parameters trained on a compilation of large Spanish unannotated corpora \cite{jose_canete_2019_3247731} and by using the Whole Word Masking technique. RoBERTa-base and RoBERTa-large are models that were developed simultaneously~\cite{maria} by the Barceloan Supercomputing Center (BSC). They are RoBERTa based models that differ mainly in the number of parameters, 115M and 774M respectively.
We fine-tune the models, using the training datasets from the section \ref{sec:trainingdata}, for five epochs on a distributed NVIDIA 3090 RTX 24GB GPUs cluster, using 24-72 GPU hours in total, depending on the size of both the model and the dataset. We used the trainer class of the Huggingface transformers \cite{wolf-etal-2020-transformers} library in python, adapted for the multiple choice paradigm with cross-entropy loss and Adam as the optimization algorithm. The learning rate was set to 2e−5 with a weight decay of 0.01 and with a batch size of 16 and kept all other hyperparameters at their default values. We perform model selection choosing the checkpoint with highest accuracy on a validation dataset based on accuracy. Note that for training we use $k=4$ senses,  including the target sense. For testing we perform as many test as necessary to reach the total number of meanings of a given target word, and the candidate sense with the highest score is the predicted sense produced by the system.

\vspace{-0.25cm}
\subsection{Evaluation}

Table \ref{table:table_all_in_one} displays the results obtained for the BERT and RoBERTa models fine-tuned using various combinations of datasets, as well as the performance of recent multilingual approaches for the Spanish WSD task in the SE13 and SE15 benchmarks.
The SDS dataset offers a remarkably significant enhancement on the SE15 benchmark over the preceding multilingual WSD systems in Spanish WSD. This improvement is consistent for both BERT and RoBERTa models. In such a case, all Spanish language models trained with a combination including the SDS dataset surpassed the Multimirror system, with the RoBERTa-large model achieving a substantial improvement of +9.5 points in F1-score. For the SE13 benchmark, only the SemCor+SDS dataset still surpasses the best multilingual WSD system but with a moderated improvement of +1.6 points in F1-score for the BETO system.  The improvement in F1-score is also observed when comparing within each Spanish language model. Specifically, the combination of the MuLaN and SC training datasets along with the SDS training set yields to an improvement of +1 point in F1-score for both SE13 and SE15. It is worth noting that the RoBERTa-large model is able to leverage the potential of combining MuLaN-SDS by reporting the best F1-score, 78.12\% on the DLEval benchmark.

\vspace{-0.25cm}
\subsection{Discussion}

Table \ref{table:table_all_in_one} shows the benefit of using a lexicon knowledge manually curated by experts and Spanish-only trained models, in comparison to currently multilingual systems for native Spanish WSD task. The best results obtained on the DLEval are those reported by the models fine-tuned employing the SDS dataset in contrast with the results for the SemEval benchmarks. A significant disparity of approximately 15-25\% was observed when comparing the models trained with SC-SDS compared to those trained using the SDS dataset, indicating a considerable mismatch between the two benchmarks that might be due to the automatic translations in the case of SE benchmarks. Finally and based on the results, it is reported that RoBERTa models consistently demonstrate superior performance compared to the BETO model. This is likely attributed to the better data curation of the datasets utilized during the pre-training and the higher number of parameters for the large model.

\vspace{-0.25cm}
\section{Conclusion}
This paper presents a novel Spanish lexicon evaluation resource, which offers an extensive coverage and encompasses a wide range of potential lexical combinations. The DLEval exhibits exceptional precision, owing to its entirely manual validation by high-level experts. 
The models' performance is capable of either matching or surpassing the state-of-the-art results achieved by most approaches for the Spanish Word Sense Disambiguation task. This demonstrates the advantages of incorporating lexical knowledge, specifically expert validated senses and glosses, in fine-tuning neural models.

\bibliographystyle{IEEEtran}
\bibliography{mybib}

\end{document}

%% file: table_semevals.tex

\begin{table}[!tp]

\tabcolsep=0.28cm
\small
\begin{NiceTabular}{c|ccccc}[colortbl-like]	
            \toprule

              Set & Instances &  WT & WAP &  IAP &  PM  \\
                \midrule
                \rowcolor{gray!15}  SE13-wn & 1260 & 541 & 4.20 & 5.52 & 421  \\
	        
    			SE15-wn & 1043 & 507 & 6.17 & 6.99 & 446  \\

                \bottomrule

\end{NiceTabular}
  {\caption{Comparison between SemEval-2013 and  SemEval-2015 in the wn split. Here it is showed the amount of Instances, Word Types(WT), Word Average Polysemy (WAP),  Instance Average Polysemy(IAP),Polysemous Words(PW).\vspace{-0.4cm}}\label{table:table_semevals}}
\end{table}

%% file: table_comparison_eval.tex

\begin{table}[!tbp]

\tabcolsep=0.14cm
\small
\begin{NiceTabular}{c|cccccc}[colortbl-like]	
            \toprule

             Set & Instances &  SW &  MW &  Entities &  MSI &   MSL 
             \tabularnewline   \midrule
			\rowcolor{gray!15}  SE13 & 1481 & 1103 & 129 & 249 & 1.15 & 1.19 \\
	        
    			SE15 & 1239 & 1088 & 67 & 84 & 6.8 & 6.8 \\
   
			\rowcolor{gray!15}  DLEval & 12269 & 12269 & 0 & 0 & 1.67 & 1.67 \\
                  SDS & 46224 & 46224 & 0 & 0 & 2.78 & 2.78 \\

                \bottomrule

\end{NiceTabular}
  {\caption{Comparison of Spanish Test Sets between SemEval-2013, SemEval-2015, DLEval and SDS. Here it is showed the amount of Instances, Single-words (SW), Multi-words (MW), Entities, Mean senses per instance (MSI), and Mean senses per lemma (MSL).}\label{table:table_comparison_eval}}
\end{table}

%% file: table_dle_example.tex
\begin{table}[!tp]
    \centering
    \tabcolsep=0.28cm
    \footnotesize
    \begin{NiceTabular}{p{0.80\linewidth}}[colortbl-like]
        \toprule
            \textbf{DLEval sentence example and meaning tags and glosses} \\
            \setlength{\parindent}{1mm} \textit{Siete \textbf{tazas} de caldo} \\
            \setlength{\parindent}{1mm} Taza\#NOUN: A183451 \textbf{A121616} \\
            \setlength{\parindent}{17.3mm} A22450 A139788\\
            A139788: Receptáculo del retrete.\\
            A121616: Cantidad que cabe en una taza. 
        \tabularnewline \midrule
             
            \textbf{XML example} \\
            
            \textless{}sentence id="d001.s10699"\textgreater \\
            
            \setlength{\parindent}{3mm}\textless{}wf lemma="siete" pos="ADJ"\textgreater{}Siete\textless{}/wf\textgreater \\
            
            \setlength{\parindent}{3mm}\textless{}instance id="d001.s10699.t0001" lemma="taza" pos="NOUN"\textgreater{}tazas\textless{}/instance\textgreater \\
            
            \setlength{\parindent}{3mm}\textless{}wf lemma="de" pos="ADP"\textgreater{}de\textless{}/wf\textgreater \\
            
             \setlength{\parindent}{3mm}\textless{}wf lemma="caldo" pos="NOUN"\textgreater{}caldo\textless{}/wf\textgreater \\
            
            \textless{}/sentence\textgreater{}\
        \tabularnewline \midrule
            \textbf{Gold file} \\
            d001.s10699.t0001 \textbf{A121616} \\
            \bottomrule
       
    \end{NiceTabular}
    \caption{DLEval sentence example and XML following the WSD standard format for the target word "taza" and four different meanings.\vspace{-0.5cm}}
    \label{table:table_dle_example}
\end{table}

%% file: table_all_in_one.tex
\begin{table}[!tbp]
\tabcolsep=0.12cm
\small
\begin{NiceTabular}{c|l|l|l|l}[colortbl-like]
\toprule

Model                      & Dataset    & SE13  & SE15  & DLEval          \\
\midrule
\midrule
mBERT-UNI \cite{su-etal-2022-multilingual}        & MuLaN         & 69.68 & 67.11 & -            \\
UKB-Graph \cite{maru-etal-2019-syntagnet}              &\rowcolor{gray!15} Syntagnet         & 73.4  & 61.2  & -            \\
AMuSE-WSD$^{*}$ \cite{orlando-etal-2021-amuse}                  & SC-WNG          & 80.0  & 73.0  & -            \\
SensEMBERT \cite{Scarlini_Pasini_Navigli_2020}                &\rowcolor{gray!15} SC-BN        & 74.6  & 64.1  & -            \\
Multimirror \cite{ijcai2021p539}                 & SC-WNG-BN       & 82.17 & 70.42 & -            \\

\midrule
     {}                      & \rowcolor{gray!15} SDS        & 63.71 & 68.72 & 73.89 \\
     {}                      & MuLaN      & 76.35 & 74.33 & 44.01 \\
    {BETO}        & \rowcolor{gray!15}SemCor(SC) & 82.06 & 77.22 & 55.48 \\
     {}                      & MuLaN-SC   & 78.44 & 74.91 & 41.29 \\
     {}                      & \rowcolor{gray!15}SC-SDS     & 83.83 & 79.00 & 73.97\\
     {}                      & MuLaN-SDS  & 74.54 & 76.13 & 73.40 \\
\midrule
                           & \rowcolor{gray!15}SDS        & 66.00 & 64.53 &    72.36          \\
                           & MuLaN      & 78.47 & 70.18 & 42.69        \\
                           & \rowcolor{gray!15}SemCor(SC) & 83.24 & 77.57 & 53.00        \\
  {RoBERTa-base}& MuLaN-SC   & \bf{83.87} & 77.95 & 48.29        \\
                           & \rowcolor{gray!15}SC-SDS     & 83.48 & 78.23 & 70.59        \\
                           & MuLaN-SDS  & 75.71 & 76.82 &  70.50            \\
\midrule  
                            & \rowcolor{gray!15}SDS        & 65.11 &   63.35    &  74.83         \\
                            & MuLaN   &   78.08    &   75.90    &   43.80           \\
                            &\rowcolor{gray!15} SemCor(SC) & 81.54 &  77.30   &    53.18            \\
{RoBERTa-large} & MuLaN-SC   &   79.50    &  \bf{80.59}     &     53.65         \\
                           &\rowcolor{gray!15} SC-SDS     &  82.45   & 78.96      &    71.11          \\
                           & MuLaN-SDS  &  79.60     &  77.52    &   \bf{78.12}         \\
\bottomrule
\end{NiceTabular}
\caption{F1-score(\%) comparison of current SOTA models for Multilingual/Spanish WSD against pre-trained Spanish models finetuned with different datasets. $^{*}$AMuSE-WSD results on full multilingual dataset.\vspace{-0.75cm}}
\label{table:table_all_in_one}
\end{table}